\documentclass{IEEEtran}
\usepackage{amsmath,amstext,amsthm,amssymb}
\usepackage{latex8}
\usepackage{times}

\usepackage{amsmath,amstext,amssymb,epsf}
\usepackage{epsfig}
\usepackage{verbatim}
\usepackage{pslatex}

\newtheorem{coro}{\sc Corollary}
\newtheorem{nota}{\sc Notation}
\newtheorem{defin}{\sc Definition}
\newtheorem{rem}{\sc Remark}
\newtheorem{cla}{\sc Claim}
\newtheorem{ex}{\sc Example}

\title{\Huge Information Distance: New Developments}
\author{Paul M.B. Vit\'anyi\\CWI, Amsterdam, The Netherlands\\
({\em Invited Lecture})\thanks{Affiliation: National Research Center for Mathematics and Computer Science in the Netherlands (CWI). 
Address: CWI, Science Park 123, 1098XG Amsterdam,
The Netherlands. Email: Paul.Vitanyi@cwi.nl}}

\begin{document}
\maketitle

\begin{abstract}
In pattern recognition, learning, and data mining one obtains
information from information-carrying objects. This involves
an objective definition of the information
in a single object, the information to go from one object to
another object in a pair of objects, the information to go from
one object to any other object in a multiple of objects,
and the shared information between objects. This is called ``information
distance.''
We survey a selection of new developments in information distance. 
\end{abstract}

\section{The Case $n=2$}
The clustering we use is hierarchical clustering in dendrograms
based on a new fast heuristic for the quartet method \cite{CV11}.
If we consider $n$ objects, then we find $n^2$ pairwise distances.
These distances are between natural data. We let the data decide for
themselves, and construct a hierarchical clustering of the $n$ objects
concerned. For details see the cited reference.
The method takes the  $n \times n$ distance matrix as input, and
yields a dendrogram with the $n$ objects as leaves (so the
dendrogram contains $n$ external nodes or leaves and $n-2$ internal nodes.
We assume $n \geq 4$.
The method is available as an open-source software tool, \cite{Ci03}.

Our aim
is to capture, in a single similarity metric,
{\em every effective distance}:
effective versions of Hamming distance, Euclidean distance,
edit distances, alignment distance, Lempel-Ziv distance,
and so on.
This metric should be so general that it works in every
domain: music, text, literature, programs, genomes, executables,
natural language determination,
equally and simultaneously.
It would be able to simultaneously detect {\em all\/}
similarities between pieces that other effective distances can detect
seperately.

Such a ``universal'' metric
was co-developed by us as a normalized
version of the ``information metric'' of \cite{BGLVZ,LV08}.
There it was shown that the information metric minorizes up to a constant all
effective distances satisfying a mild density requirement (excluding
for example distances that are 1 for every pair $x,y$ such that $x \neq y$).
This justifies the notion that the information distance is universal.

We may be interested what happens
in terms of properties or features of the 
pair of objects analyzed, say $x$ and $y$.
It can be shown that the information distance captures every property 
of which the Kolmogorov complexity
is logarithmic in the length of $\min\{|x|,|y|\}$. If those lengths go to
infinity, then logarithm of those lengths go to infinity too.
In this case the information distance captures every property. 

This information distance (actually a metric up to minor additive terms)
is normalized so that the resulting
distances are in $[0,1]$ and can be shown to retain the metric
property, \cite{Li02}.
The result is the 
``normalized information distance'' (actually a metric up to neglidgible terms).
All this is in terms of Kolmogorov complexity \cite{LV08}.
 
It articulates the intuition that two objects are deemed close if
we can significantly ``compress'' one given the information
in the other, that is, two pieces are more similar
if we can more succinctly describe one given the other.
The normalized information distance
 discovers all effective similarities in the sense that if two
objects are close according to some effective similarity, then 
they are also close according to the normalized information distance.

Put differently, the normalized information distance represents
similarity according to the dominating shared feature between
the two objects being compared.
In comparisons of more than two objects, 
different pairs may have different dominating features.
For every two objects,
this normalized information metric distance zooms in on the dominant
similarity between those two objects
 out of a wide class of admissible similarity
features. 
Since the normalized information distance also satisfies the metric 
(in)equalities, and takes values in $[0,1]$,
it may be called {\em ``the'' similarity metric}.

Unfortunately, the universality of the normalized information distance
comes at the price of noncomputability. Recently we have shown 
that the normalized 
information distance is not even semicomputable (this is weaker than
computable) and there is no semicomputable function at a computable
distance of it \cite{TTV11}.

Since the Kolmogorov
complexity of a string or file is the length 
of the ultimate compressed version of that
file, 
we can use real data compression programs to approximate the Kolmogorov
complexity.
Therefore, to apply this ideal precise mathematical theory in real life,
we have to replace the use of  the noncomputable
Kolmogorov complexity by an approximation
using a standard real-world compressor. 
Starting from the normalized information distance, 
if $Z$ is a compressor and we use $Z(x)$
to denote the length of the compressed version of a string $x$,
then we arrive at the {\em Normalized Compression Distance}:
\begin{equation}\label{eq.ncd}
 NCD(x,y) = \frac{Z(xy) - \min(Z(x),Z(y))}{\max(Z(x),Z(y))},
\end{equation}
where for convenience we have replaced the pair $(x,y)$ in the formula
by the concatenation $xy$, and we ignore logarithmic terms
in the numerator and denominator,
see \cite{Li02,CV04}.
In \cite{CV04} we propose axioms to capture the real-world setting,
and show that \eqref{eq.ncd}
approximates optimality.
Actually, the
NCD is a family of compression functions parameterized 
by the given data
compressor $Z$. 

\subsection{Web-based Similarity}
To make computers more intelligent one would like
to represent meaning in computer-digestable form.
Long-term and labor-intensive efforts like
the {\em Cyc} project \cite{Le95} and the {\em WordNet}
project \cite{Miea} try to establish semantic relations
between common objects, or, more precisely, {\em names} for those
objects. The idea is to create
a semantic web of such vast proportions that rudimentary intelligence
and knowledge about the real world spontaneously emerges.
This comes at the great cost of designing structures capable
of manipulating knowledge, and entering high
quality contents in these structures
by knowledgeable human experts. While the efforts are long-running
and large scale, the overall information entered is minute compared
to what is available on the Internet.

The rise of the Internet has enticed millions of users
to type in trillions of characters to create billions of web pages of
on average low quality contents. The sheer mass of the information
available about almost every conceivable topic makes it likely
that extremes will cancel and the majority or average is meaningful
in a low-quality approximate sense. Below, we give a general
method to tap the amorphous low-grade knowledge available for free
on the Internet, typed in by local users aiming at personal
gratification of diverse objectives, and yet globally achieving
what is effectively the largest semantic electronic database in the world.
Moreover, this database is available for all by using any search engine
that can return aggregate page-count estimates like Google for a large
range of search-queries.

While the previous NCD method that compares the objects themselves using
\eqref{eq.ncd} is
particularly suited to obtain knowledge about the similarity of
objects themselves, irrespective of common beliefs about such
similarities, we now develop a method that uses only the name
of an object and obtains knowledge about the similarity of objects
by tapping available information generated by multitudes of
web users.
The new method is useful to extract knowledge from a given corpus of
knowledge, in this case the pages on the Internet  accessed by a search engine
returning aggregate page counts, but not to
obtain true facts that are not common knowledge in that database.
For example, common viewpoints on the creation myths in different
religions
may be extracted by the web-based method, but contentious questions
of fact concerning the phylogeny of species can be better approached
by using the genomes of these species, rather than by opinion.
This approach was proposed by \cite{CV07}. We skip the theory.

In contrast to strings $x$ where the complexity $Z(x)$ represents
the length of the compressed version of $x$ using compressor $Z$, for a search
term $x$ (just the name for an object rather than the object itself),
the  code of length $G(x)$ represents the shortest expected
prefix-code word length of the event ${\bf x}$ (the number of pages
of the Internet returned by a given search engine).
The associated 
{\em normalized web distance} (NWD) is defined
just as \eqref{eq.ncd} with the search engine in the role of compressor
yielding code lengths $G(x), G(y)$
for the singleton search terms $x,y$ being compaired
and a code length $G(x,y)$ for the doubleton pair $(x,y)$, by
\begin{equation}\label{eq.NGD}
 NWD(x,y)=\frac{G(x,y) - \min(G(x),G(y))}{\max(G(x),G(y))}.
\end{equation}
This $NWD$ 
uses the background knowledge on the web as viewed by the search engine
as conditional information. 

The 
same formula as \eqref{eq.NGD} can be written
in terms of frequencies of the number of pages returned on a search
query as
\begin{equation}\label{eq.ngd}
NWD(x,y) = \frac{  \max \{\log f(x), \log f(y)\}  - \log f(x,y) \}}{
\log N - \min\{\log f(x), \log f(y) \}},
\end{equation}
and if $f(x),f(y)>0$ and $f(x,y)=0$ then $NWD(x,y)= \infty$.
It is easy to see that
\begin{enumerate}
\item
$NWD(x,y)$ is undefined for  $f(x)=f(y)=0$;
\item
$NWD(x,y) = \infty$ for $f(x,y)=0$ and either or both $f(x)>0$
and $f(y)>0$; and
\item
$ NWD(x,y) \geq 0$ otherwise.
\end{enumerate}
The number $N$ is related to the number of pages $M$ indexed by the search
engine we use. Our experimental results suggest that every reasonable
(greater than any $f(x)$) value can be used for the normalizing factor  $N$,
and our
results seem  in general insensitive to this choice.  In our software, this
parameter $N$ can be adjusted as appropriate, and we often use $M$ for $N$.
In the \cite{CV07} we analyze the mathematical properties of NWD,
and  prove the universality of the search engine distribution.
We show that the NWD is not a metric, in contrast to the NCD.
The generic example showing the nonmetricity of semantics (and therefore
the NWD) 
is that a man is close to a centaur,
and a centaur is close to a horse, but a man is very different from
a horse.

\subsection{Question-Answer System}
A typical procedure for finding an answer on the Internet consists in
entering some terms regarding the question into a Web search engine and then
browsing the search results in search for the answer. This is particularly
inconvenient when one uses a mobile device with a slow internet connection and
small display.  Question-answer (QA) systems attempt to solve this problem.
They allow the user to enter a question in natural language and generate an
answer by searching the Web autonomously.
the QA system
QUANTA \cite{ZHZL08} that uses variants of the NCD and the NWD to
identify the correct answer to a question out of several candidates for
answers. QUANTA is remarkable in that it uses neither NCD nor NWD
introduced so far, but a variation that is nevertheless based on the same
theoretical principles. This variation is tuned to the particular needs of a QA
system. Without going in too much detail it uses the maximal overlap
of program $p$ going from file $x$ to file $y$, and program $q$ going
from file $y$ to file $x$. The system QUANTA is 1.5 times better
(according to generally used measures) than its competition.

\section{$n > 2$}
In many applications we are interested in
shared information between {\em many} objects
instead of just a pair of objects. For example, in customer reviews of
gadgets, in blogs about public happenings,
in newspaper articles about the same occurrence, we are interested
in the most comprehensive one or the most specialized one.
Thus, we want to extend the information distance
measure from pairs to multiples. This approach was introduced in \cite{Li08}
while most of the theory is developed in \cite{Vi11}.

Let $X$ denote a finite list of
$m$ finite binary strings defined by $X=(x_1, \ldots, x_m)$, the constituting
strings ordered length-increasing lexicographic.
We use lists and not sets, since if $X$ is a set we cannot express
simply the distance from a string to itself or between strings that
are all equal.
Let $U$ be the reference universal Turing machine.
Given the string $x_i$ we define the information distance to any string in $X$
by $E_{\max} (X) = \min \{|p|: U(x_i,p,j)=x_j$ for all $x_i,x_j \in X$\}.
It is shown in \cite{Li08}, Theorem 2, that
\begin{equation}\label{eq.li08}
E_{\max}(X)=
\max_{x:x \in X} K(X|x),
\end{equation}
up to a logarithmic additive term.
Define $E_{\min}(X) = \min_{x:x \in X} K(X|x)$.
Theorem 3 in \cite{Li08} states that for every list $X= (x_1, \ldots , x_m)$ we have
\begin{equation}\label{eq.li083}
E_{\min}(X) \leq E_{\max}(X) \leq
\min_{i: 1 \leq i \leq m} 
\sum_{x_i,x_k \in X \; \& \; k \neq i} E_{\max} (x_i,x_k),  
\end{equation}
up to a logarithmic additive term. This is not a corollary of
\eqref{eq.li08} as stated in \cite{Li08}, but
both inequalities follow from the definitions.
The lefthand side is interpreted as
the program length of the
``most comprehensive object that contains the most information
about all the others [all elements of $X$],''
and the righthand side is interpreted as the program length
of the ``most specialized object that
is similar to all the others.''

Information distance for multiples, that is, finite lists, appears
both practically and theoretically
promising. The results below appear in \cite{Vi11}.
In all cases the results imply the corresponding
ones for the pairwise
information distance defined as follows.
The information distance in \cite{BGLVZ}
between strings $x_1$ and $x_2$ is $E_{\max}(x_1,x_2) 
= \max\{K(x_1|x_2),K(x_2|x_1)\}$.
In the \cite{Vi11} $E_{\max}(X) = \max_{x:x \in X} K(X|x)$.
These two definitions coincide for $|X|=2$ since $K(x,y|x)=K(y|x)$ up
to an additive constant term.
The reference investigate
the maximal overlap of information 
which for $|X|=2$ specializes to Theorem 3.4 in \cite{BGLVZ}.
A corollary in \cite{Vi11} shows \eqref{eq.li08} and another corollary
shows that the lefthand side of \eqref{eq.li083} can indeed be taken to
correspond to a single program embodying the ``most comprehensive object
that contains the most information about all the others'' as stated but not
argued or proved in \cite{Li08}. The reference proves
metricity 
and universality 
which for $|XY|=2$ (for metricity) and $|X|=2$ (for universality) specialize to
Theorem 4.2 in \cite{BGLVZ};
additivity;
minimum overlap of information 
which for $|X|=2$ specializes to Theorem 8.3.7 in \cite{Mu02};
and the nonmetricity of
normalized information distance
for lists of more than two elements and the failure of certain proposals
of a normalizing factor (to achieve a normalized version).
In contrast, for
lists of two elements we can normalize the
information distance as in Lemma V.4 and Theorem V.7 of
\cite{Li02}.
The definitions are of necessity new as are the proof ideas.
Remarkably, the new notation and proofs for the general case
are simpler than the mentioned existing proofs for the particular case
of pairwise information distance.

\section{Conclusion}
\label{sect.exp}
By now applications abound. 
See the many references to the papers
\cite{Li02,CV04,CV07} in Google Scholar. 

The methods turns out to be more-or-less 
robust under change of the underlying
compressor-types: statistical (PPMZ), Lempel-Ziv based  dictionary (gzip),
block based (bzip2), or special purpose (Gencompress). Obviously the window
size matters, as well as how good the compressor is. For example, PPMZ
gives for mtDNA of the investigated species diagonal elements ($NCD(x,x)$)
between 0.002 and 0.006. The compressor bzip2 does considerably worse,
and gzip gives something in between 0.5 and 1 on the diagonal elements.
Nonetheless, for texts like books gzip does fine in our experiments;
the window size is sufficient and we do not use the diagonal elements.
But for genomics gzip is no good.


\begin{thebibliography}{99}


\bibitem{BGLVZ}
C.H. Bennett, P. G\'acs, M. Li, P.M.B. Vit\'anyi, W. Zurek,
Information Distance, {\em IEEE Trans. Information Theory},
44:4(1998), 1407--1423.









\bibitem{Ci03}
R.L. Cilibrasi, The CompLearn Toolkit, 2003--,
 www.complearn.org





\bibitem{CV04}
R.L. Cilibrasi, P.M.B. Vit\'anyi, Clustering by compression, 
{\em IEEE Trans. Information Theory}, 51:4(2005), 1523--1545.


\bibitem{CV07}
R.L. Cilibrasi, P.M.B. Vit\'anyi,
{\em IEEE Trans. Knowledge and Data Engineering}, 19:3(2007), 370--383.

\bibitem{CV11}
R.L. Cilibrasi, P.M.B. Vit\'anyi, A fast quartet tree heuristic for 
hierarchical clustering, {\em Pattern Recognition}, 44 (2011) 662-677






\bibitem{Ko65}
A.N. Kolmogorov,
Three approaches to the quantitative definition of information,
{\em Problems Inform. Transmission}, 1:1(1965), 1--7.

\bibitem{Le95}
D.B. Lenat,
Cyc: A large-scale investment in knowledge infrastructure,
{\em Comm. ACM}, 38:11(1995),33--38.



\bibitem{Li02}
M. Li, X. Chen, X. Li, B. Ma, P.M.B. Vit\'anyi.
The similarity metric,
{\em IEEE Trans. Information Theory}, 50:12(2004), 3250- 3264.

\bibitem{LV08}
M. Li, P.M.B. Vit\'anyi.
{\em An Introduction to Kolmogorov Complexity and Its Applications},
3nd Ed.,
Springer-Verlag, New York, 2008.

\bibitem{Li08}
M. Li, C. Long, B. Ma, X. Zhu, Information shared by many objects,
Proc. 17th ACM Conf. Inform. Knowl. Management,
2008, 1213--1220.



\bibitem{Miea}
G.A. Miller et.al, WordNet, 
A Lexical Database for the English Language,
Cognitive Science Lab, Princeton University.

\bibitem{Mu02}
An.A. Muchnik, Conditional complexity and codes,
{\em Theor. Comput. Sci.}, 271(2002), 97--109.



\bibitem{TTV11}
S.A. Terwijn, L. Torenvliet, P.M.B. Vit\'anyi, 
Nonapproximability of the Normalized Information Distance, 
{\em J. Comput. System Sciences}, 77:4(2011), 738--742. 

\bibitem{Vi11}
P.M.B. Vit\'anyi, Information distance in multiples, 
{\em IEEE Trans. Inform. Theory}, 57:4(2011), 2451-2456.


\bibitem{ZHZL08}
X. Zhang, Y. Hao, X.-Y. Zhu, M. Li,
New Information Distance Measure and Its Application in Question
Answering System, {\em J. Comput. Sci. Techn.}, 23:4(2008), 557--572.

\end{thebibliography}
\end{document}